\documentclass[letterpaper, 10 pt, conference]{ieeeconf}  

\IEEEoverridecommandlockouts                              

\usepackage{xcolor}
\usepackage{graphicx}
\usepackage{amsmath}
\usepackage{tikz}
\usepackage{hyperref}
\usepackage{gensymb}

\usepackage{multirow}
\usepackage{url}
\usepackage{flushend}
\usepackage{balance}

\title{\LARGE \bf Designing Kresling Origami for Personalised Wrist Orthosis*}

\author{Chenying Liu$^{1,2}$, Shuai Mao$^{1}$, Yixing Lei$^{1,2}$, and Liang He$^{1,2}$
\thanks{*This work is funded by the Podium Institute for Sports Medicine and Technology at the University of Oxford.}
\thanks{$^{1}$Chenying Liu, Shuai Mao, Yixing Lei, and Liang He are with the Healthcare Biorobotics Lab, Institute of Biomedical Engineering, Department of Engineering Science, University of Oxford, Oxford OX1 3PJ, United Kingdom. All inquiries can be addressed to:
    {\tt\small chenying.liu@eng.ox.ac.uk}.}%
\thanks{$^{2}$Chenying Liu, Yixing Lei, and Liang He are also with the Podium Institute for Sports Medicine and Technology, University of Oxford, Oxford OX3 7DQ, United Kingdom.}
}

\begin{document}

\maketitle
\thispagestyle{empty}
\pagestyle{empty}

\begin{abstract}

The wrist plays a pivotal role in facilitating motion dexterity and hand functions. Wrist orthoses, from passive braces to active exoskeletons, provide an effective solution for the assistance and rehabilitation of motor abilities. However, the type of motions facilitated by currently available orthoses is limited, with little emphasis on personalised design. To address these gaps, this paper proposes a novel wrist orthosis design inspired by the Kresling origami. The design can be adapted to accommodate various individual shape parameters, which benefits from the topological variations and intrinsic compliance of origami. Heat-sealable fabrics are used to replicate the non-rigid nature of the Kresling origami. The orthosis is capable of six distinct motion modes with a detachable tendon-based actuation system. Experimental characterisation of the workspace has been conducted by activating tendons individually. The maximum bending angle in each direction ranges from 18.81° to 32.63°. When tendons are pulled in combination, the maximum bending angles in the dorsal, palmar, radial, and ulnar directions are 31.66°, 30.38°, 27.14°, and 14.92°, respectively. The capability to generate complex motions such as the dart-throwing motion and circumduction has also been experimentally validated. The work presents a promising foundation for the development of personalised wrist orthoses for training and rehabilitation.

\end{abstract}

\section{INTRODUCTION}

Wrist impairments like muscle weakness can significantly impact an individual's quality of life and daily functioning \cite{mcglinchey2020effect}. For example, the immobility of the wrist makes everyday tasks, such as gripping objects or writing, more challenging or even impossible without assistance \cite{adams2003impact}. Therefore, physical therapy and rehabilitation are crucial for assisting people with wrist impairments to rebuild their dexterity and strength. In the past, occupational interventions have been used in combination with video guides and physiotherapy to assist with wrist motion rehabilitation \cite{roll2017effectiveness}. Most recently, wrist exoskeletons have reached clinical stages and demonstrated promising results in rebuilding motor functions \cite{charles2005wrist, pitzalis2023state1}. 

\begin{figure}
    \centering
    \includegraphics[width=\linewidth]{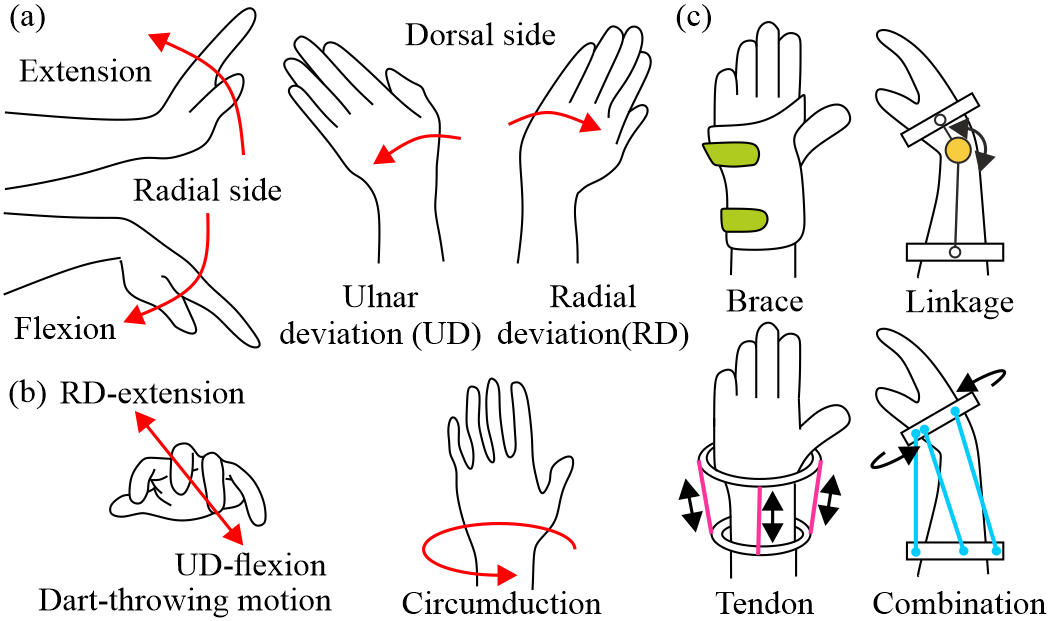}
    \caption{Wrist motions and orthoses. (a) Basic movements. (b) Complex movements. (c) Working principles of wrist braces and robotic exoskeletons.}
    \label{wrist motion}
\end{figure}

In general, the wrist joint is capable of four basic movements \cite{youm1980kinematics}, namely extension, flexion, radial deviation and ulnar deviation (see Fig. \ref{wrist motion}(a)). It should be noted that movements enabled by the forearm, pronation and supination, are not included in this analysis. Furthermore, the wrist joint can also produce complex movements such as the dart-throwing motion (DTM) and circumduction as illustrated in Fig. \ref{wrist motion}(b). In particular, DTM denotes the transition from radial deviation-extension to ulnar deviation-flexion \cite{wolfe2006dart}. The circumduction movement involves a circular trajectory of the wrist joint which connects all basic movements \cite{salvia2000analysis}. All these wrist movements can be decomposed into bending-like motions and transitions between different bent states.

A range of wrist orthoses, from passive braces to active exoskeletons, have been developed to assist with wrist joint rehabilitation \cite{tan2020soft}. For instance, the brace depicted in Fig. \ref{wrist motion}(c) or splint is a passive device to constrain the wrist movement \cite{koo2017development}. Robotic exoskeletons are active devices comprising linkages, tendons, or a combination of both to replicate the movements of the wrist. These exoskeletons can be attached to the wrist limbs to assist individuals who have reduced muscle strength \cite{esmaeili2013hyperstaticity}. The working principles of these devices vary. For example, the linkage system has a rotational joint with one to three degrees of freedom (DoF), and can generate basic wrist movements \cite{amoozandeh2022design, pezent2017design}. Some tendon-based designs come with cables that can be pulled to reduce the distance between the palm and forearm, which enables the wrist to move \cite{gerez2019development}. Furthermore, length-variable actuators, such as pneumatic artificial muscles, are also a tendon to realise similar motions with enhanced compliance \cite{andrikopoulos2015design}. The linkage and tendon designs allow the wrist to bend in multiple directions to mimic basic movements and DTM \cite{choi2019exo}. The combined system depicted in Fig. \ref{wrist motion}(c) is a parallel mechanism similar to the Stewart platform, which has been designed specifically for circumduction movements \cite{kitano2018development}.

Despite considerable advances in research and commercial domains, the majority of wrist orthoses can only accommodate basic wrist motions with limited versatility to be extended for complex movements \cite{pitzalis2023state2}. The underlying reason is that most existing devices are designed as a low-DoF linkage or made from soft materials. Consequently, these devices may lack the required dexterity or controllability to facilitate a greater variety of wrist movements. Additionally, there has been a very limited emphasis on the varying wrist and palm sizes of different individuals. A personalised design methodology is hence needed.

\begin{figure}
    \centering
    \includegraphics[width=\linewidth]{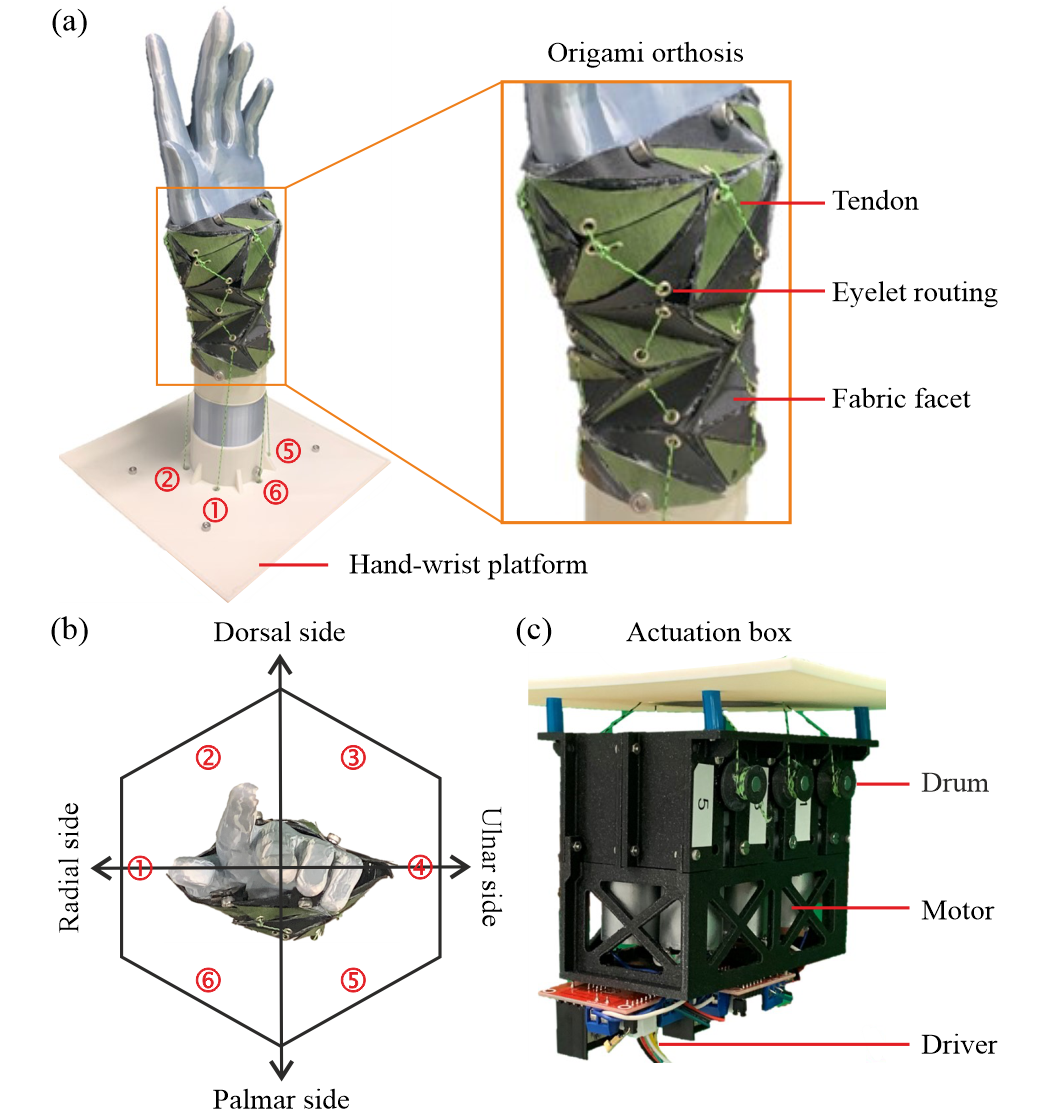}    
    \caption{Kresling origami-inspired wrist orthosis. (a) Orthosis structure on a hand-wrist platform. (b) Tendon arrangement. (c) Actuation system.}
    \label{wrist orthosis demo}
\end{figure}

In recent years, origami, known as the paper folding technique, has emerged as a promising solution to offer diverse motion capabilities and variable design parameters. \cite{liu2024exploring, liu20213d}. This feature makes it possible to develop a personalised wrist orthosis from origami that can facilitate versatile wrist movements. Notably, the concept of origami has already been adopted in wrist orthosis design. For instance, a variation of Yoshimura origami was used to design modular actuators to drive basic wrist movements \cite{liu2021compact}. The same origami was also geometrically optimised to serve as the skeleton of an orthosis designed to immobilise the limb \cite{barros2022computational}. However, none of these designs has successfully achieved a personalised fit while simultaneously enabling versatile wrist movements.

Our work introduces a novel wrist orthosis as illustrated in Fig. \ref{wrist orthosis demo}(a). The key structure of this orthosis draws inspiration from the Kresling origami with omnidirectional bending motions \cite{wu2021stretchable, kaufmann2022harnessing}. The proposed design leverages the topological variants of the Kresling origami and its inherent compliance to facilitate custom fit. Heat-sealable fabric materials have been used in the manufacturing process to imitate the non-rigid nature of origami, where the facets can be bent or stretched during the folding process. Finally, the wrist orthosis can be equipped with a tendon-driven actuation system shown in Figs. \ref{wrist orthosis demo}(b) and \ref{wrist orthosis demo}(c), which supports both basic and complex wrist movements.

The remainder of this paper is organised as follows. Section \MakeUppercase{\romannumeral 2} illustrates the design, working principle, and manufacturing process of the wrist orthosis. Section \MakeUppercase{\romannumeral 3} outlines the experimental setups to assess the wrist orthosis' performance. Results and discussions are presented in Section \MakeUppercase{\romannumeral 4}. Finally, we conclude with Section \MakeUppercase{\romannumeral 5}.

\section{WRIST ORTHOSIS DESIGN}

\subsection{Kresling Origami and Its Conical Variation}

\begin{figure}
    \centering
    \includegraphics[width=\linewidth]{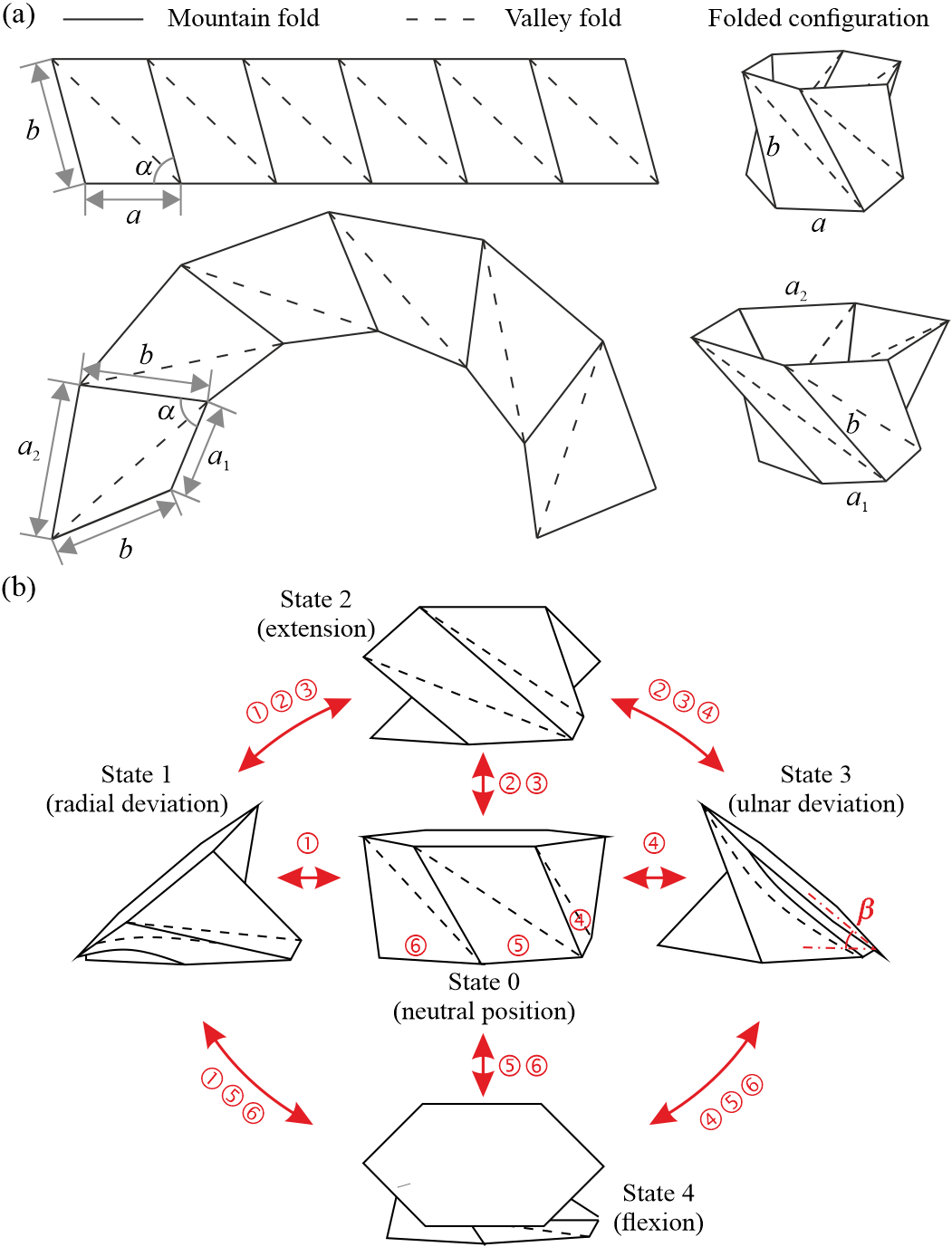}
    \caption{Kresling origami and its bending motions. (a) Crease patterns of traditional Kresling origami, conical Kresling origami, and their folded configurations. (b) Bending motions of a semi-folded Kresling origami. The bending angle $\beta$ is roughly indicated for State 3.}
    \label{Kresling origami}
\end{figure}

\begin{figure*}
    \centering
    \includegraphics[width=\linewidth]{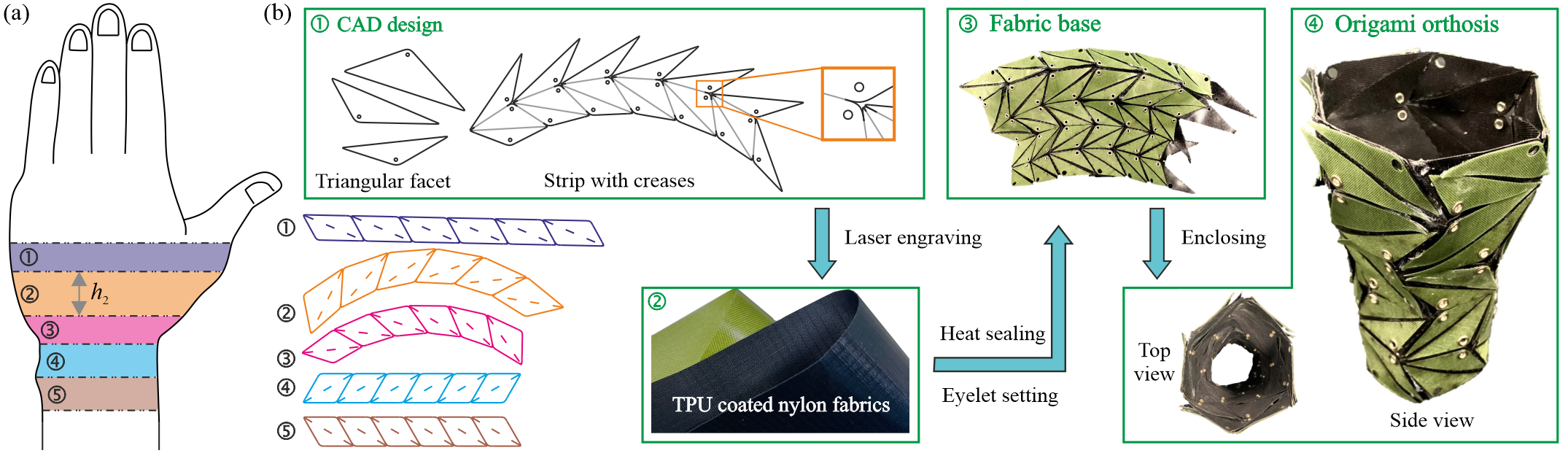}
    \caption{Design and fabrication of origami wrist orthosis. (a) Topological design to fit hand-wrist-forearm characteristics. (b) Detailed manufacturing steps.}
    \label{orthosis topology}
\end{figure*}

The crease patterns and folded configurations of the Kresling origami are illustrated in Fig. \ref{Kresling origami}(a). Each pattern consists of six cells, which are then connected in a loop to form a unit with two regular hexagons parallel to each other at the top and bottom. The upper pattern represents the traditional Kresling origami (TKO) proposed by Biruta Kresling \cite{kresling2020fifth}. Each cell is a parallelogram, and thus the two hexagons are identical. The lower pattern represents the conical Kresling origami (CKO), which has been extensively studied by Lu et. al \cite{lu2022conical}. The cell is a quadrilateral with a pair of opposite sides of equal length. The strip exhibits a curved shape rather than a straight arrangement. This geometric feature gives hexagons of two sizes at the top and bottom. While it is not possible to connect the TKO and CKO strips on the same plane, they can be aligned in space provided that the following geometric condition is met.

\begin{equation}
a=a_1\;\text{or}\;a=a_2
\end{equation}

The connected creases are mountain folds, which were originally hexagonal sides. Accordingly, this method allows for the arrangement and connection of multiple TKO and CKO units in a tubular configuration, which forms the backbone of the wrist orthosis.

\subsection{Working Principle Enabled by Omnidirectional Bending}

The Kresling origami can bend omnidirectionally (i.e., in any direction) from its semi-folded neutral position. To illustrate the working principle, a single TKO unit is taken as an example. As depicted in Fig. \ref{Kresling origami}(b), the semi-folded unit can bend from State 0 to States 1–4, mimicking four basic wrist movements. The unit can also bend in any direction and transit between different states, thus enabling complex wrist movements. It should be noted that the triangular facets on the side are bent or stretched in the bending process while the top and bottom hexagons remain unaltered. However, the hexagons are no longer parallel and the angle between them is defined as the bending angle $\beta$ as shown in Fig. \ref{Kresling origami}(b). 

Six tendons, T\textsubscript{1} - T\textsubscript{6}, are used to facilitate the bending process as illustrated in Fig. \ref{wrist orthosis demo}(a). For each cell of a unit, a tendon originates from the upper right corner of the parallelogram and terminates diagonally at the bottom left vertex. A tensile force applied to the tendon will result in a reduction in the distance between the points of attachment, which closes the diagonal crease. These creases are designed elastic, allowing the structure to return to its original configuration when the tendons are released. 

The tendon arrangement is shown in Fig. \ref{wrist orthosis demo}(b). The tendons can be pulled (P) or loosened (L) individually or in combination to imitate a specific wrist movement. A summary of the tendon states for wrist movements is in Table \ref{tendon arrangement}. The key action tendon (KAT) refers to the simplest tendon arrangement that must be pulled to realise a specific movement. Yet, alternative tendon arrangements can produce a similar function. For example, pulling T\textsubscript{3} - T\textsubscript{5} results in ulnar extension similar to that achieved by pulling T\textsubscript{4}.

\begin{table}[h]
\caption{Wrist Movements and Tendon States}
\label{tendon arrangement}
\renewcommand{\arraystretch}{1.2}
\begin{center}
\begin{tabular}{|c|c|c|c|c|c|c|c|}
\hline
\textbf{Tendon}               & \textbf{1}   & \textbf{2}   & \textbf{3}   & \textbf{4}   & \textbf{5}   & \textbf{6}   & \textbf{KAT}  \\ \hline
\textbf{Extension}            & L   & P   & P   & L   & L   & L   & 2, 3 \\ \hline
\textbf{Flexion}             & L   & L   & L   & L   & P   & P   & 5, 6 \\ \hline
\textbf{Radial deviation}     & P   & P   & L   & L   & L   & P   & 1    \\ \hline
\textbf{Ulnar deviation}      & L   & L   & P   & P   & P   & L   & 4    \\ \hline
\multirow{2}{*}{\textbf{DTM}} & P   & P   & P   & L   & L   & L   & 2    \\ \cline{2-8} 
                     & L   & L   & L   & P   & P   & P   & 5    \\ \hline
\textbf{Circumduction}        & P/L & P/L & P/L & P/L & P/L & P/L & All  \\ \hline
\end{tabular}
\end{center}
\end{table}

It should be noted that circumduction involves a gradual bending in all directions, either clockwise or anticlockwise, with each state being switched to the adjacent one. To illustrate, the process may be initiated by pulling T\textsubscript{1}, which is then combined with T\textsubscript{2}. T\textsubscript{1} is subsequently loosened, leaving T\textsubscript{2} alone, after which T\textsubscript{3} is added, and so on.

The omnidirectional bending feature is preserved in the CKO unit. When the top and bottom hexagons are in contact and the facets opposite to the bending direction are fully stretched, the bending angle $\beta$ also attains its maximum value, which can be estimated using the following equations.

For States 1 and 3,
\begin{equation}
\beta=\arccos{\frac{3a_1^2+4a_2^2-b^2\sin^2{\alpha}}{4\sqrt{3}a_1a_2}}
\label{radial and ulnar}
\end{equation}

For States 2 and 4,
\begin{equation}
\beta=\arccos{\frac{4a_1^2+3a_2^2-b^2\sin^2{\alpha}}{4\sqrt{3}a_1a_2}}
\label{dorsal and palmer}
\end{equation}
where $a_1=a_2=a$ for TKO.

\subsection{Topological Design of a Custom Fit Orthosis}

The wrist orthosis covers the palm, wrist, and the anterior aspect of the forearm as highlighted in Fig. \ref{orthosis topology}(a). These areas are divided into five sections, each of which will be enveloped by a layer of a Kresling origami unit. The upper edge of Section 1 and the lower edge of Section 5 are affixed to the palm and forearm, respectively.

Given the geometric parameters of the hand-wrist-forearm model, Sections 1, 4, and 5 are intended to be enclosed by TKO units. Sections 3 and 4 exhibit considerable variation between the top and bottom circumferences, and thus will be addressed by CKO units. For Section $i$, the top and bottom circumferences are measured as $c_{i.t}$ and $c_{i.b}$, respectively, with the distance between them designated as $d_i$. The parameters of each unit are provided below.

For Sections 1, 4, and 5 where $c_{i.t}=c_{i.b}$,
\begin{equation}
    a_{i1}=a_{i2}=\frac{c_{i.t}+t_i}{6}
\end{equation}

For Sections 2 and 3 where $c_{i.t}\neq c_{i.b}$,
\begin{equation}
    a_{i1}=\frac{c_{i.b}+t_i}{6}\;\text{and}\;a_{i2}=\frac{c_{i.t}+t_i}{6}
\end{equation}
where $t_i$ represents the circumference tolerance and accounts for shrinkage on the inner side of the Kresling origami unit.

Furthermore, to ease the fabrication and operation of wrist orthosis, $\alpha$ has been set to $60^\circ$ and $t_i$ is taken as 15 mm for all units. For Section 1, we have $b_1=0.6a_{11}$.  In Sections 2 – 5, the value of $b_i$ is initially set equal to that of $a_{i1}$. Let $h_i$ indicates the height of each section as exemplified in Fig. \ref{orthosis topology}(a). $b_i$ must meet the condition $b_i\sin{\alpha}>h_i$ to guarantee that each unit can be in a semi-folded configuration to enable bending. Otherwise, it is recommended that the value of $b_i$ should be increased by $20\%$ until it reaches the minimum requirement. The final crease patterns for all sections are given in Fig. \ref{orthosis topology}, where adjacent patterns are in reverse directions to minimise the twisting motion from the Kresling origami.

\subsection{Manufacturing Process}

The orthosis is fabricated using thermoplastic polyurethane (TPU) coated nylon fabrics, which are suitable for replicating the non-rigid nature of the Kresling origami. The manufacturing process is illustrated in Fig. \ref{orthosis topology}(b). The computer-aided design (CAD) of crease patterns and facets was used to laser engrave fabrics. Notably, foldable creases (grey lines) were realised through dash line cutting and vertices were modified to rounded shapes to avoid physical interference due to material thickness. Small apertures were created to accommodate eyelets for tendon routing. Additionally, extra materials were retained on the strip to facilitate the formation of a loop or interconnection with adjacent strips. The strip and triangular facets were subsequently affixed with TPU coatings in contact. An iron set at 200°C was used to heat the materials for one minute until both layers were fully bonded. This process was then repeated on the adjacent strips, which were then connected to form a fabric base. The base was subsequently heated once more to create a closed loop.

Braided fishing lines (Dorisea, UK) were taken as tendons. They were attached to the orthosis via eyelet routing as illustrated in Fig. \ref{wrist orthosis demo}(a). For each tendon, one end was affixed to Section 1 with the other coming out from Section 5. Sections 2 - 4 would thus account for the majority of bending motions. An actuation box was developed as shown in Fig. \ref{wrist orthosis demo}(c) to manipulate six tendons. The system was composed of six high-torque gear motors, three L298N drivers, a microcontroller (Arduino UNO), and a power supply.

\section{EXPERIMENTS}

\subsection{Selection of Heat-Sealable Fabrics}

Three types (70D, 210D, and 420D) of heat-sealable TPU-coated nylon are available from IRON RAFT, UK. The ultimate tensile strengths of these materials were measured on the INSTRON machine, and the one with a relatively higher value was selected to ensure sufficient rigidity of facets and elasticity of foldable creases. The three types of fabrics exhibit tensile strengths of 7.44 MPa, 15.59 MPa, and 16.83 Mpa, respectively. Hence, type 420D was chosen.

\subsection{Wrist Model and Orthosis Fabrication Parameter}

Two physical hand-wrist models were used to evaluate the performance of the wrist orthosis. Each wrist joint was a semi-spherical joint covered by an elastic membrane. The geometry of each model and the associated orthosis design parameters are provided in Table \ref{design parameter}. Model 1 features a smaller hand and a thinner wrist, whereas Model 2 is larger. An unfit orthosis was also fabricated as the control group. 

\begin{table}[h]
\caption{Wrist Model Specifics and Origami Design Parameters}
\label{design parameter}
\renewcommand{\arraystretch}{1.2}
\begin{center}
\begin{tabular}{|c|c|c|c|c|c|}
\hline
\textbf{Model 1}    & $c_{1.t}$  & $c_{1.b}$/$c_{2.t}$ & $c_{2.b}$/$c_{3.t}$ & $c_{3.b}$/$c_{4.t}$ & $c_{4.b}$/$c_{5.t}$ \\ \hline
Unit: mm   & 225.3 & 225.3     & 185.1     & 153.9    & 153.9     \\ \hline
\textbf{Model 1}    & $h_1$  & $h_2$ & $h_3$ & $h_4$ & $h_5$ \\ \hline
Unit: mm   & 18 & 27     & 21     & 20    & 22     \\ \hline
\textbf{Orthosis 1} & $a_{12}$    & $a_{11}$/$a_{22}$       & $a_{21}$/$a_{32}$   & $a_{31}$/$a_{42}$   & $a_{41}$/$a_{52}$        \\ \hline
Unit: mm   & 40.0  & 40.0      & 33.0      & 28.0     & 28.0      \\ \hline
\textbf{Model 2}    & $c_{1.t}$  & $c_{1.b}$/$c_{2.t}$ & $c_{2.b}$/$c_{3.t}$ & $c_{3.b}$/$c_{4.t}$ & $c_{4.b}$/$c_{5.t}$ \\ \hline
Unit: mm   & 258.2 & 258.2     & 213.5     & 175.6    & 175.6     \\ \hline
\textbf{Model 2}    & $h_1$  & $h_2$ & $h_3$ & $h_4$ & $h_5$ \\ \hline
Unit: mm   & 20 & 29     & 24     & 26    & 26     \\ \hline
\textbf{Orthosis 2} & $a_{12}$    & $a_{11}$/$a_{22}$       & $a_{21}$/$a_{32}$   & $a_{31}$/$a_{42}$   & $a_{41}$/$a_{52}$        \\ \hline
Unit: mm   & 45.7  & 45.7      & 37.7      & 32.0     & 32.0      \\ \hline
\textbf{Unfit} & $a_{12}$    & $a_{11}$/$a_{22}$       & $a_{21}$/$a_{32}$   & $a_{31}$/$a_{42}$   & $a_{41}$/$a_{52}$        \\ \hline
Unit: mm   & 40.0  & 40.0      & 40.0      & 40.0     & 40.0      \\ \hline
\end{tabular}
\end{center}
\end{table}

\begin{figure}
    \centering
    \includegraphics[width=\linewidth]{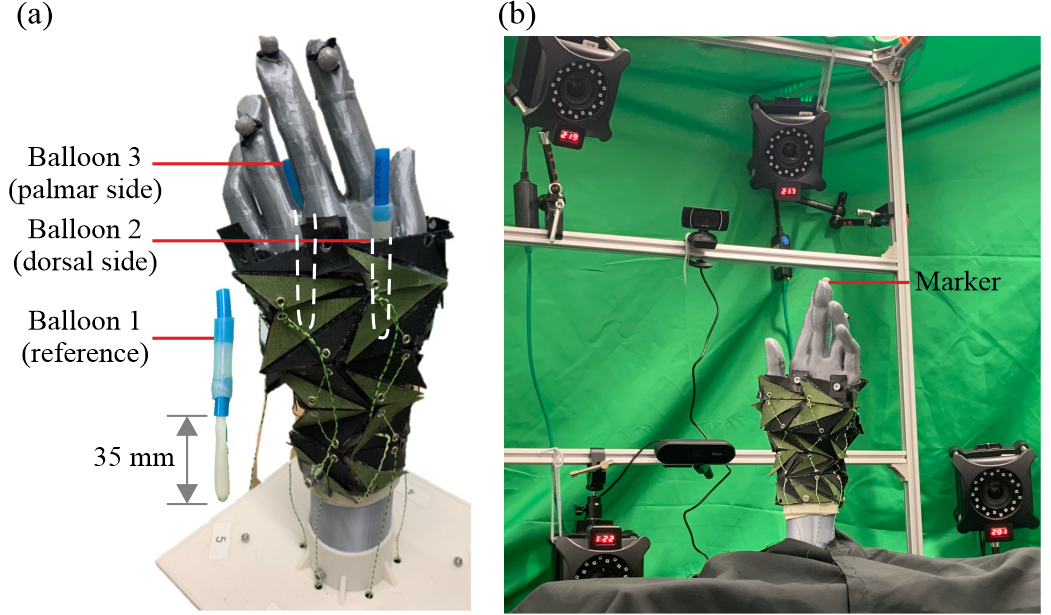}
    \caption{Experimental setups. (a) Balloons for custom fit evaluation. (b) Actuated wrist orthosis in the motion tracking system for workspace and movement characterisation.}
    \label{experiment setup}
\end{figure}

\begin{figure*}
    \centering
    \includegraphics[width=\linewidth]{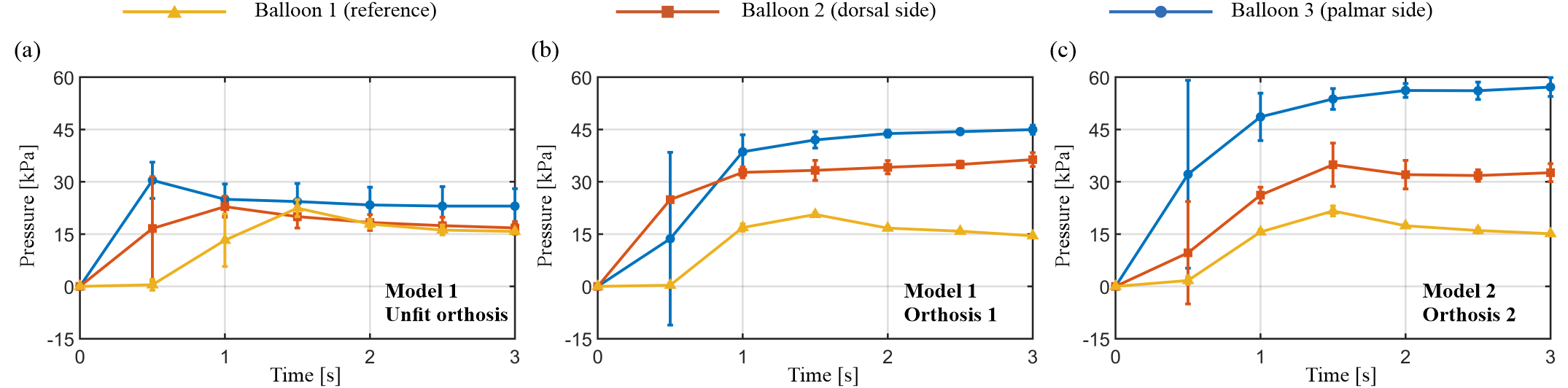}
    \caption{Internal pressure of balloons. (1) Unfit wrist orthosis on Model 1. (2) and (3) Personalised wrist orthoses on Model 1 and Model 2, respectively. }
    \label{balloon pressure}
\end{figure*}

\subsection{Evaluation of Customised Fit}

Three balloons were used to assess the level of fit of wrist orthoses. Balloon 1 was left in the air as a reference, while the other two were positioned on the dorsal and palmar sides of each hand-wrist model as shown in Fig. \ref{experiment setup}(a). The balloons were inflated at a constant flow rate for 3 seconds, and their internal pressures were recorded at 2 Hz. Three groups of experiments were conducted, i.e., Model 1 with Orthosis 1, Model 2 with Orthosis 2, and Model 1 with the unfit orthosis. Each experiment was repeated 3 times.

\subsection{Characterisation of Workspace and Wrist Movement}

Model 2 with Orthosis 2 was placed under a motion tracking system, MARS4H (NOKOV Ltd, China) to characterise its workspace and specific movements as shown in \ref{experiment setup}(b). Three markers were attached to the index, middle, and ring fingers, whose positions were recorded. The following motions were conducted and repeated 3 times each.

\subsubsection{Workspace}Each tendon was subjected to a 7-second pulling and 7-second loosening sequence. This was done to let the wrist bend towards each hexagonal side and then return to the neutral position.

\subsubsection{Basic movement}Set T\textsubscript{2} and T\textsubscript{3}, Set T\textsubscript{5} and T\textsubscript{6}, T\textsubscript{1}, and T\textsubscript{4}, were pulled and loosened in sequence for 7 seconds, respectively. This was done following the methodology outlined in Fig. \ref{Kresling origami}(b) and Table \ref{tendon arrangement}, which were designed to enable extension, flexion, ulnar, and radial deviations. Here, only the KAT was engaged without exhausting all possible tendon arrangements. The same principle applied to the dart-throwing motion and circumduction described below.

\subsubsection{Dart-throwing motion}T\textsubscript{2} was pulled and then loosened for 7 seconds each, followed by the activation of T\textsubscript{5} for the same motion. This was done to facilitate the transition from radial deviation-extension to ulnar deviation-flexion. Furthermore, the mirror motion, namely the transition between ulnar deviation-extension and radial deviation-flexion, was investigated by activating T\textsubscript{3} and T\textsubscript{6} in the same pattern.

\subsubsection{Circumduction}T\textsubscript{1} was pulled for 7 seconds, after which it was loosened for 14 seconds. T\textsubscript{2} was activated in the same way, beginning at a point 3.5 seconds after the procedure started. T\textsubscript{3} was initiated when T\textsubscript{1} began to be loosened. T\textsubscript{4} - T\textsubscript{6} were then actuated in sequence following the same strategy. This was to enable the orthosis to transit from one bent state to its adjacent one without passing through the neutral position.

A video demonstrating workspace characterisation and wrist movement validation is included in the Appendix.

\section{RESULTS AND DISCUSSIONS}

\subsection{Customised Fit Level of Wrist Orthosis}

The internal pressure of each balloon is plotted in Fig. \ref{balloon pressure} with its standard deviation. The pressure data were averaged from 3 repeated experiments. After 2 seconds, the reference balloon 1 reaches a steady pressure of approximately 15 kPa. For the unfit orthosis, a slight increase in pressure is observed in balloons 2 and 3. The increase in pressure is more obvious in Orthoses 2 and 3, which were designed for a personalised fit. The observed pressure increase can be attributed to the gap between the hand-wrist model and orthosis, which is not inflatable. Thus, a bigger pressure increase indicates a smaller gap, and hence an improved fit design.

Additionally, the data reported in Fig. \ref{balloon pressure} can be used to infer the pressure exerted by the orthosis on the hand-wrist model. In practice, human tactile proprioception has a pressure-pain threshold that varies between individuals \cite{ozcan2004comparison}. Therefore, this data serves as an indicator to prevent excessive pressure that could cause discomfort or pain. When testing the orthosis on human subjects, this method can be employed to refine the orthosis parameters, ensuring adequate customisation so that the device is neither too tight nor too loose.

\subsection{Workspace}

\begin{figure*}
    \centering
    \includegraphics[width=\linewidth]{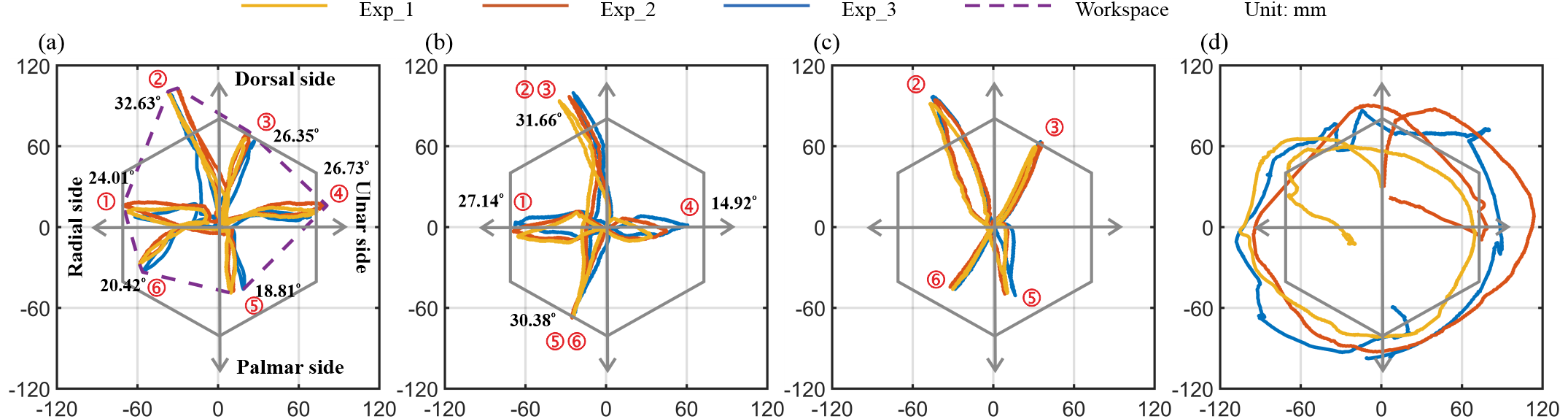}
    \caption{Workspace and movements of the origami orthosis. (a) Workspace by activating each tendon individually. (b) Basic movements, namely extension, flexion, radial, and ulnar deviations. (c) Dart-throwing and its mirror motions. (d) Circumduction motion.}
    \label{motion trajectory}
\end{figure*}

The trajectory of the middle finger was selected to present the orthosis workspace. The spatial trajectory is projected onto a plane, as illustrated in Fig. \ref{motion trajectory}(a). The results exhibit good repeatability. It is noteworthy that the trajectory of the middle finger is not evenly distributed. Specifically, the angles between adjacent trajectories are not equal, which in theory should be $60^\circ$. Additionally, bending angles in all directions are averaged from 3 experiments. The maximum bending angle is $32.63^\circ$ facilitated by T\textsubscript{2}, whilst the lowest angle is $18.81^\circ$  enabled by T\textsubscript{5}. This discrepancy can be attributed, at least in part, to the mismatch between the regular hexagonal shape of the wrist orthosis and the oval cross-section of the palm. The compliance of the fabric base may also have played a role.

Additionally, a deviation in the trajectory of the finger is observed when the tendons are pulled and loosened. This hysteresis may be attributed to the fact that the force on each tendon is not perfectly perpendicular to the folding crease. Consequently, the actual folding line might be shifted by a minor angle. When the tendon is released, the crease is unfolded along its original position.

\subsection{Basic Movements}

The wrist orthosis has demonstrated four basic movements, as illustrated in Fig. \ref{motion trajectory}(b). The radial and ulnar deviations are aligned with the theoretical directions, while the extension and flexion tendons exhibit a radial inclination. The bending angle along each direction is calculated, with the values averaged from the 3 experiments. According to a biomechanical study, the normal functional range of wrist motion for daily tasks is $5^\circ$ of flexion, $30^\circ$ of extension, $10^\circ$ of radial deviation, and $15^\circ$ of ulnar deviation \cite{palmer1985functional}. Hence, the bending range of the proposed orthosis is sufficient to support most everyday activities.

Sections 1 and 5 were affixed to the palm and forearm, respectively, and their geometric design was intended solely to ensure a customised fit without contributing to any movement. Therefore, the deformation of the orthosis should be concentrated in Sections 2 - 4. According to Eqs. \ref{radial and ulnar} and \ref{dorsal and palmer}, along with Table \ref{design parameter}, these three sections yield theoretical maximum bending angles of $57.21^\circ$ in the radial/ulnar direction and $66.84^\circ$ in the dorsal/palmar direction. These bending angles are significantly larger than the experimental data. The reasons for this are analysed as follows.

It has been observed that, in the experiments, only Sections 3 and 4 - those closest to the wrist joint - undergo significant deformation, while Section 2 remains nearly unchanged on the palm and does not contribute to the bending process. Theoretically, the maximum bending angles of Section 3 along radial/ulnar and dorsal/palmar directions are $17.06^\circ$ and $24.72^\circ$, respectively. For Section 4, the bending angle is up to $25.56^\circ$. In the dorsal, palmar, and radial directions, the observed angles exceed the maximums of the individual sections but are less than their sum. Comparing the experimental data with these theoretical values, we can deduce that both sections contribute to the bending process, although none reach their maximum bending angle. In contrast, the ulnar direction exhibits a relatively small bending angle, which may be attributed to the lack of sufficient folding of the facets on that side.

\subsection{Complex Movements}

DTM, its mirror motion, and circumduction have been validated as shown in Figs. \ref{motion trajectory}(c) and \ref{motion trajectory}(d). In the case of DTM and its mirror motion, tendons were activated individually like that of workspace characterisation. However, the trajectories are not identical to that of Fig. \ref{motion trajectory}(d). The sequence of tendon actuation has exerted a considerable influence. This is also reflected in the circumduction motion, which exhibits a slightly larger and more evenly distributed workspace than the original one. Therefore, we hypothesise that the presence of a deformed adjacent facet facilitates bending movements.

\subsection{Limitations and Future Work}

Currently, the fabrication process relies on manual heat-sealing, which, while tedious, is still effective for creating the entire origami orthosis. This is particularly important due to the non-developable nature of the structure, which means it cannot be unfolded into a flat sheet without stretching \cite{zou2024kinematics}. In the future, 3D printing TPU directly onto fabric materials could help automate the process; however, its feasibility in addressing non-developable patterns needs to be explored.

In terms of experimental characterisation, it is important to note that not all possible tendon arrangements were evaluated to analyse the orthosis movements. While it is expected that arrangements other than the KAT could produce similar movements as explained before, they may result in different motion trajectories. Further experiments involving all tendon arrangements, combined with alternative tendon routings, are needed to fully understand behaviours of the orthosis and unlock its real-world potential.

Last but not least, the compliance of the fabric orthosis has been observed to influence its workspace and motion modes. Additionally, discrepancies in the trajectories of the same mode can be attributed to geometric mismatches and tendon-actuation dynamics. These observations highlight opportunities for further refinement through a more comprehensive modelling approach that incorporates material compliance, tendon interactions, and geometric mismatches. Such refinements are essential for advancing to human subject testing and tailoring the orthosis to individual needs.

\section{CONCLUSIONS}

This paper presents a novel wrist orthosis design inspired by the Kresling origami. A tendon-based actuation system has been developed to facilitate bending motions similar to the human wrist. The topological arrangement of the Kresling origami allows the orthosis to be designed to fit different wrist attributes with simple measurements. The actuated orthosis has been validated on a hand-wrist model to achieve basic and complex wrist movements. The workspace and bending angles of the orthosis have been experimentally characterised. This work represents a crucial first step to personalised orthosis design.

\addtolength{\textheight}{-12cm}   




\section*{APPENDIX}

\label{appendix}

The orthosis workspace and wrist movement can be viewed at: \url{https://youtu.be/BDod0H_ubJE}. 


\section*{ACKNOWLEDGMENT}

The authors would like to thank Professor Barbara Rossi and Mr Xiongfeng Ruan's support to characterise the ultimate tensile strength of fabric materials.


\bibliographystyle{IEEETran}

\begin{thebibliography}{10}
\providecommand{\url}[1]{#1}
\csname url@rmstyle\endcsname
\providecommand{\newblock}{\relax}
\providecommand{\bibinfo}[2]{#2}
\providecommand\BIBentrySTDinterwordspacing{\spaceskip=0pt\relax}
\providecommand\BIBentryALTinterwordstretchfactor{4}
\providecommand\BIBentryALTinterwordspacing{\spaceskip=\fontdimen2\font plus
\BIBentryALTinterwordstretchfactor\fontdimen3\font minus \fontdimen4\font\relax}
\providecommand\BIBforeignlanguage[2]{{%
\expandafter\ifx\csname l@#1\endcsname\relax
\typeout{** WARNING: IEEEtran.bst: No hyphenation pattern has been}%
\typeout{** loaded for the language `#1'. Using the pattern for}%
\typeout{** the default language instead.}%
\else
\language=\csname l@#1\endcsname
\fi
#2}}

\bibitem{mcglinchey2020effect}
M.~P. McGlinchey, J.~James, C.~McKevitt, A.~Douiri, and C.~Sackley, ``The effect of rehabilitation interventions on physical function and immobility-related complications in severe stroke: a systematic review,'' \emph{BMJ open}, vol.~10, no.~2, 2020.

\bibitem{adams2003impact}
B.~D. Adams, N.~M. Grosland, D.~M. Murphy, and M.~McCullough, ``Impact of impaired wrist motion on hand and upper-extremity performance,'' \emph{The Journal of hand surgery}, vol.~28, no.~6, pp. 898--903, 2003.

\bibitem{roll2017effectiveness}
S.~C. Roll and M.~E. Hardison, ``Effectiveness of occupational therapy interventions for adults with musculoskeletal conditions of the forearm, wrist, and hand: a systematic review,'' \emph{The American journal of occupational therapy}, vol.~71, no.~1, 2017.

\bibitem{charles2005wrist}
S.~K. Charles, H.~I. Krebs, B.~T. Volpe, D.~Lynch, and N.~Hogan, ``Wrist rehabilitation following stroke: initial clinical results,'' in \emph{9th International Conference on Rehabilitation Robotics, 2005. ICORR 2005.}\hskip 1em plus 0.5em minus 0.4em\relax IEEE, 2005, pp. 13--16.

\bibitem{pitzalis2023state1}
R.~F. Pitzalis, D.~Park, D.~G. Caldwell, G.~Berselli, and J.~Ortiz, ``State of the art in wearable wrist exoskeletons part i: Background needs and design requirements,'' \emph{Machines}, vol.~11, no.~4, p. 458, 2023.

\bibitem{youm1980kinematics}
Y.~Youm and A.~E. FLATT, ``Kinematics of the wrist.'' \emph{Clinical Orthopaedics and Related Research (1976-2007)}, vol. 149, pp. 21--32, 1980.

\bibitem{wolfe2006dart}
S.~W. Wolfe, J.~J. Crisco, C.~M. Orr, and M.~W. Marzke, ``The dart-throwing motion of the wrist: is it unique to humans?'' \emph{The Journal of hand surgery}, vol.~31, no.~9, pp. 1429--1437, 2006.

\bibitem{salvia2000analysis}
P.~Salvia, L.~Woestyn, J.~H. David, V.~Feipel, S.~Van, S.~Jan, P.~Klein, and M.~Rooze, ``Analysis of helical axes, pivot and envelope in active wrist circumduction,'' \emph{Clinical Biomechanics}, vol.~15, no.~2, pp. 103--111, 2000.

\bibitem{tan2020soft}
X.~Tan, L.~He, J.~Cao, W.~Chen, and T.~Nanayakkara, ``A soft pressure sensor skin for hand and wrist orthoses,'' \emph{IEEE Robotics and Automation Letters}, vol.~5, no.~2, pp. 2192--2199, 2020.

\bibitem{koo2017development}
D.~S. Koo and J.~R. Lee, ``The development of a wrist brace using 3d scanner and 3d printer,'' \emph{Fashion \& Textile Research Journal}, vol.~19, no.~3, pp. 312--319, 2017.

\bibitem{esmaeili2013hyperstaticity}
M.~Esmaeili, N.~Jarrass{\'e}, W.~Dailey, E.~B{\"u}rdet, and D.~Campolo, ``Hyperstaticity for ergonomie design of a wrist exoskeleton,'' in \emph{2013 IEEE 13th International Conference on Rehabilitation Robotics (ICORR)}.\hskip 1em plus 0.5em minus 0.4em\relax IEEE, 2013, pp. 1--6.

\bibitem{amoozandeh2022design}
A.~Amoozandeh~Nobaveh, G.~Radaelli, and J.~L. Herder, ``A design tool for passive wrist support,'' in \emph{Wearable Robotics: Challenges and Trends: Proceedings of the 5th International Symposium on Wearable Robotics, WeRob2020, and of WearRAcon Europe 2020, October 13--16, 2020}.\hskip 1em plus 0.5em minus 0.4em\relax Springer, 2022, pp. 13--17.

\bibitem{pezent2017design}
E.~Pezent, C.~G. Rose, A.~D. Deshpande, and M.~K. O'Malley, ``Design and characterization of the openwrist: A robotic wrist exoskeleton for coordinated hand-wrist rehabilitation,'' in \emph{2017 international conference on rehabilitation robotics (ICORR)}.\hskip 1em plus 0.5em minus 0.4em\relax IEEE, 2017, pp. 720--725.

\bibitem{gerez2019development}
L.~Gerez, J.~Chen, and M.~Liarokapis, ``On the development of adaptive, tendon-driven, wearable exo-gloves for grasping capabilities enhancement,'' \emph{IEEE Robotics and Automation letters}, vol.~4, no.~2, pp. 422--429, 2019.

\bibitem{andrikopoulos2015design}
G.~Andrikopoulos, G.~Nikolakopoulos, and S.~Manesis, ``Design and development of an exoskeletal wrist prototype via pneumatic artificial muscles,'' \emph{Meccanica}, vol.~50, pp. 2709--2730, 2015.

\bibitem{choi2019exo}
H.~Choi, B.~B. Kang, B.-K. Jung, and K.-J. Cho, ``Exo-wrist: A soft tendon-driven wrist-wearable robot with active anchor for dart-throwing motion in hemiplegic patients,'' \emph{IEEE Robotics and Automation Letters}, vol.~4, no.~4, pp. 4499--4506, 2019.

\bibitem{kitano2018development}
Y.~Kitano, T.~Tanzawa, and K.~Yokota, ``Development of wearable rehabilitation device using parallel link mechanism: rehabilitation of compound motion combining palmar/dorsi flexion and radial/ulnar deviation,'' \emph{Robomech Journal}, vol.~5, pp. 1--8, 2018.

\bibitem{pitzalis2023state2}
R.~F. Pitzalis, D.~Park, D.~G. Caldwell, G.~Berselli, and J.~Ortiz, ``State of the art in wearable wrist exoskeletons part ii: A review of commercial and research devices,'' \emph{Machines}, vol.~12, no.~1, p.~21, 2023.

\bibitem{liu2024exploring}
C.~Liu, L.~He, S.~Wang, A.~Williams, Z.~You, and P.~Maiolino, ``Exploring kinematic bifurcations and hinge compliance for in-hand manipulation: How could thick-panel origami contribute?'' \emph{Advanced Intelligent Systems}, 2024.

\bibitem{liu20213d}
C.~Liu, P.~Maiolino, and Z.~You, ``A 3d-printable robotic gripper based on thick panel origami,'' \emph{Frontiers in Robotics and AI}, vol.~8, 2021.

\bibitem{liu2021compact}
S.~Liu, Z.~Fang, J.~Liu, K.~Tang, J.~Luo, J.~Yi, X.~Hu, and Z.~Wang, ``A compact soft robotic wrist brace with origami actuators,'' \emph{Frontiers in Robotics and AI}, vol.~8, p. 614623, 2021.

\bibitem{barros2022computational}
M.~O. Barros, A.~Walker, and T.~Stankovi{\'c}, ``Computational design of an additively manufactured origami-based hand orthosis,'' \emph{Proceedings of the Design Society}, vol.~2, pp. 1231--1242, 2022.

\bibitem{wu2021stretchable}
S.~Wu, Q.~Ze, J.~Dai, N.~Udipi, G.~H. Paulino, and R.~Zhao, ``Stretchable origami robotic arm with omnidirectional bending and twisting,'' \emph{Proceedings of the National Academy of Sciences}, vol. 118, no.~36, 2021.

\bibitem{kaufmann2022harnessing}
J.~Kaufmann, P.~Bhovad, and S.~Li, ``Harnessing the multistability of kresling origami for reconfigurable articulation in soft robotic arms,'' \emph{Soft Robotics}, vol.~9, no.~2, pp. 212--223, 2022.

\bibitem{kresling2020fifth}
B.~Kresling, ``The fifth fold: Complex symmetries in kresling-origami patterns,'' \emph{Symmetry: Culture and Science}, vol.~31, no.~4, pp. 403--416, 2020.

\bibitem{lu2022conical}
L.~Lu, X.~Dang, F.~Feng, P.~Lv, and H.~Duan, ``Conical kresling origami and its applications to curvature and energy programming,'' \emph{Proceedings of the Royal Society A}, vol. 478, no. 2257, 2022.

\bibitem{ozcan2004comparison}
A.~{\"O}zcan, Z.~Tulum, L.~P{\i}nar, and F.~Ba{\c{s}}kurt, ``Comparison of pressure pain threshold, grip strength, dexterity and touch pressure of dominant and non-dominant hands within and between right-and left-handed subjects,'' \emph{Journal of Korean medical science}, vol.~19, no.~6, pp. 874--878, 2004.

\bibitem{palmer1985functional}
A.~K. Palmer, F.~W. Werner, D.~Murphy, and R.~Glisson, ``Functional wrist motion: a biomechanical study,'' \emph{Journal of Hand Surgery}, vol.~10, no.~1, pp. 39--46, 1985.

\bibitem{zou2024kinematics}
Y.~Zou, F.~Feng, K.~Liu, P.~Lv, and H.~Duan, ``Kinematics and dynamics of non-developable origami,'' \emph{Proceedings of the Royal Society A}, vol. 480, no. 2282, p. 20230610, 2024.

\end{thebibliography}

\end{document}